\pdfoutput=1

\documentclass[11pt]{article}

\usepackage[]{emnlp2021}

\usepackage{times}
\usepackage{latexsym}

\usepackage[T1]{fontenc}

\usepackage[utf8]{inputenc}

\usepackage{microtype}

\usepackage{graphicx}
\usepackage{multirow}
\usepackage{amsmath,amsfonts,amssymb}
\usepackage{algorithmic}
\usepackage[ruled,vlined]{algorithm2e}
\DeclareMathOperator*{\argmax}{arg\,max}

\usepackage{bm}
\usepackage{array}
\usepackage{CJKutf8}
\usepackage{color}

\newcommand{\cmtt}[1]{{\fontfamily{cmtt}\selectfont {#1}}}

\DeclareFontShape{OT1}{cmtt}{bx}{n}
{
<5> <6> <7> <8> <9>
<10> <10.95> <12> <14.4> <17.28> <20.74> <24.88> cmbtt10
}{}
\DeclareFontShape{OT1}{cmtt}{b}{n}
{<->sub * cmtt/bx/n}{}

\title{Compositional Generalization via Semantic Tagging}

\author{Hao Zheng \textnormal{and} Mirella Lapata\\
Institute for Language, Cognition and Computation\\
School of Informatics, University of Edinburgh\\
 10 Crichton Street, Edinburgh EH8 9AB\\
\texttt{Hao.Zheng@ed.ac.uk}~~~~\texttt{mlap@inf.ed.ac.uk}\\
}

\date{}

\makeatletter
\newcommand{\thickhline}{%
    \noalign {\ifnum 0=`}\fi \hrule height 1pt
    \futurelet \reserved@a \@xhline
}
\makeatother

\begin{document}
\maketitle
\begin{abstract}
 Although neural sequence-to-sequence models have been successfully
  applied to semantic parsing, they fail at \emph{compositional
    generalization}, i.e.,~they are unable to systematically
  generalize to unseen compositions of seen components. Motivated by
  traditional semantic parsing where compositionality is explicitly
  accounted for by symbolic grammars, we propose a new decoding
  framework that preserves the expressivity and generality of
  sequence-to-sequence models while featuring lexicon-style alignments
  and disentangled information processing. Specifically, we decompose
  decoding into two phases where an input utterance is first tagged
  with semantic symbols representing the meaning of individual words,
  and then a sequence-to-sequence model is used to predict the final
  meaning representation conditioning on the utterance \emph{and} the
  predicted tag sequence. Experimental results on three semantic
  parsing datasets show that the proposed approach consistently
  improves compositional generalization across model architectures,
  domains, and semantic formalisms.\footnote{Our code and data can be
    found at \url{https://github.com/mswellhao/Semantic-Tagging}.}
\end{abstract}

\section{Introduction}

\begin{table}[t]
\centering
\begin{tabular}{p{6.2cm}} \hline
\multicolumn{1}{c}{\em Training Set} \\ \hline
\multicolumn{1}{c}{What is the density of Texas?}\\
\cmtt{select density from state where state\_name =
  "texas"} \\\hline
\multicolumn{1}{c}{\em Test Set (Question split)} \\ \hline
\multicolumn{1}{c}{What is the population density of Maine?} \\
\cmtt{select density from state where state\_name =
  "maine"} \\ \hline
\multicolumn{1}{c}{\em Test Set (Query Split)} \\ \hline
\multicolumn{1}{c}{How many people live in Washington?} \\ 
\cmtt{select population from state where state\_name =
 "washington"} \\ \hline
\end{tabular}
\caption{Two test examples from the
  question- and query-based splits of \textsc{GeoQuery} and a training example included in both splits. The  example in
  the question-based split shares the same query pattern as the
  training example while the  example in the query-based split has
  a query pattern different from the training example.} 
\label{fig:split}
\vspace{-1.5ex}

\end{table}


Semantic parsing aims at mapping natural language utterances to
machine-interpretable meaning representations such as executable
queries or logical forms. Sequence-to-sequence neural networks
\cite{SutskeverVL14} have emerged as a general modeling framework for
semantic parsing, achieving impressive results across different
domains and semantic formalisms
(\citealt{dong-lapata-2016-language,jia-liang-2016-data,iyer-etal-2017-learning,wang-etal-2020-rat},
\emph{inter alia}). Despite recent success, there has been mounting
evidence
\cite{finegan-dollak-etal-2018-improving,keysers2020measuring,herzig2020spanbased,DBLP:conf/icml/LakeB18}
that these models fail at \emph{compositional generalization},
i.e,~they are unable to systematically generalize to \emph{unseen}
compositions of \emph{seen} components. For example, a model that
observed at training time the questions ``\textsl{How many people live
  in California?}''  and ``\textsl{How many people live in the capital
  of Georgia?}'' fails to generalize to questions such as ``\emph{How
  many people live in the capital of California?}''. This is in stark
contrast with human language learners who are able to systematically
generalize to such compositions \cite{Fodor:Pylyshyn:1988,Lake2019HumanFL}.

Previous work \cite{finegan-dollak-etal-2018-improving} has exposed
the inability of semantic parsers to generalize compositionally simply
by evaluating their performance on different dataset splits. Existing
datasets commonly adopt \emph{question-based} splits where many
examples in the test set have the same query templates (induced by
anonymizing named entities) as examples in the training. As a result,
many of the queries in the test set are seen in training, and parsers
are being evaluated for their ability to generalize to questions with
different surface forms but the same meaning. In contrast, when
adopting a \emph{query-based} split, the structure of the queries in
the test set is unobserved at training time, and parsers therefore
must generalize to questions with different
meanings. Table~\ref{fig:split} illustrates the difference between
question- and query-based splits on
\textsc{GeoQuery} \cite{Zelle1996}.


On the contrary, compositional generalization poses no problem for
traditional semantic parsers \cite{Luke2005,
zettlemoyer-collins-2007-online, wong-mooney-2006-learning,
wong-mooney-2007-learning,liang-etal-2013-learning} which typically
use a (probabilistic) grammar; the latter defines the meaning of
individual words and phrases and how to best combine them in order to
obtain meaning representations for entire utterances. Neural semantic
parsers do away with representing symbolic structure explicitly in
favor of a more general approach which directly transduces the
utterance into a logical form, avoiding domain-specific assumptions
and grammar learning.

Nonetheless, the symbolic paradigm provides two important insights
that could serve as a guide in designing neural semantic parsers with
better compositional generalization.  Firstly, the probability of a
logical form is decomposed into \emph{local} factors under strong
conditional independence assumptions while in neural semantic parsing
the prediction of each symbol directly depends on \emph{all}
previously decoded symbols. This strong expressivity may hurt
compositional generalization since different kinds of information are
bundled together, rendering the model's predictions susceptible to
irrelevant context changes. Secondly, there exist \emph{hard}
alignments between logical constructs and linguistic expressions but
in neural parsers the two are only loosely related via the \emph{soft}
attention mechanism. Explicit alignments can help distinguish which
language segments are helpful for predicting certain components in the
logical form, potentially improving compositional generalization.

In this paper, we devise a new decoding framework that preserves the
expressivity and generality of sequence-to-sequence models while
featuring lexicon-style alignments and disentangled information
processing.  Specifically, we decompose decoding into two phases.
Given a natural language utterance, each word is first labeled with a
semantic symbol representing its meaning via a tagger. Semantic
symbols are atomic units like predicates (in
\mbox{$\lambda$-calculus}) or columns (in SQL). The tagger
\emph{explicitly} aligns semantic symbols to tokens or token spans in
the utterance. Moreover, the prediction of each semantic symbol is
conditionally independent of other symbols in the logical form. This
is reminiscent of lexicons in classical semantic parsers, but a major
difference is that our tagger is a neural model which considers
information based on the entire utterance and can generalize to new
words. A sequence-to-sequence model takes the utterance and predicted
tag sequence which serves as a soft constraint on the output space,
and generates the final meaning representation. Our framework is
general in that it could incorporate any sequence-to-sequence model as
the base model and augment it with semantic tagging.

We evaluate the proposed approach on query-based splits of three
semantic parsing benchmarks: \textsc{Atis}, \textsc{GeoQuery}, and a
subset of \textsc{WikiSQL} covering different semantic formalisms
(\mbox{$\lambda$-calculus} and SQL). We report experiments with LSTM-
and Transformer-based models \cite{dong-lapata-2016-language,
dong-lapata-2018-coarse, NIPS2017_7181} demonstrating that our
framework improves compositional generation across datasets and model
architectures.  Our approach is also superior to a recent data
augmentation proposal \cite{andreas-2020-good}, specifically designed
to enhance compositional generalization.

\section{Related Work}
\label{sec:related-work}

The realization that neural sequence models perform poorly in settings
requiring compositional generalization has led to several research
efforts aiming to study the extent of this problem and how to handle
it. For instance, recent studies have proposed benchmarks which allow
to measure different aspects of compositional
generalization. 

\citet{DBLP:conf/icml/LakeB18} introduce \textsc{SCAN}, a grounded
navigation task where a learner must translate natural language
commands into a sequence of actions in a synthetic
language. \citet{bahdanau2018systematic} use a synthetic VQA task to
evaluate whether models can reason about all possible object pairs
after training only on a small subset. They show that modular
structured models are best in terms of systematic generalization,
while end-to-end versions do not generalize as
well. \citet{keysers2020measuring} introduce a method to
systematically construct benchmarks for evaluating compositional
generalization. Using Freebase as an example, they create questions
which maximize compound divergence (e.g., combinations of entities and
relations) while guaranteeing that the atoms (aka the primitive
elements used to compose these questions) remain the same between
train and test sets.

Other work proposes data augmentation as a way of injecting a
compositional inductive bias into neural sequence models. Under this
protocol, synthetic examples are constructed by taking real training
examples and replacing (possibly discontinuous) fragments with other
fragments that appear in at least one similar
environment. Recombination operations can be performed by applying
rules \cite{andreas-2020-good} or learned using a generative model
\cite{Akyurek:ea:2020}.  \citet{herzig2020spanbased} follow a more
traditional approach \cite{Zelle1996, ge-mooney-2005-statistical,
  Luke2005, wong-mooney-2006-learning, wong-mooney-2007-learning,
  zettlemoyer-collins-2007-online, kwiatkowksi-etal-2010-inducing,
  kwiatkowski-etal-2011-lexical} and develop a span-based parser which
predicts a  tree over an input utterance, explicitly encoding how
partial programs compose over spans in the input. Finally,
\citet{Oren:ea:2020} improve compositional
generalization with the use of contextual representations, extensions
to decoder attention, and downsampling examples from frequent
templates.


We decompose decoding in two stages where the input is first tagged with semantic symbols which are then subsequently used to predict the final meaning representation. These semantic tags are automatically induced from logical forms without any extra annotation and vary depending on the meaning representation at hand (e.g., $\lambda$-calculus, SQL). They serve the goal of injecting inductive bias for compositional generalization rather than expressing general semantic information across languages (see \citealt{abzianidze-bos-2017-towards} for a proposal to develop a universal semantic tagset for non-executable semantic parsing). 
Our framework can be applied to different sequence-to-sequence models,
domains, and semantic formalisms. It does not require manual
task-specific engineering \cite{herzig2020spanbased} and is orthogonal
to data augmentation methods \cite{andreas-2020-good,Akyurek:ea:2020}
and other extensions \cite{Oren:ea:2020} which we could also
incorporate.



\section{Model Architecture}
\label{sec:model}

\begin{figure}[t]
\centering
\hspace*{-.2cm}\hbox{\includegraphics[width=.49\textwidth]{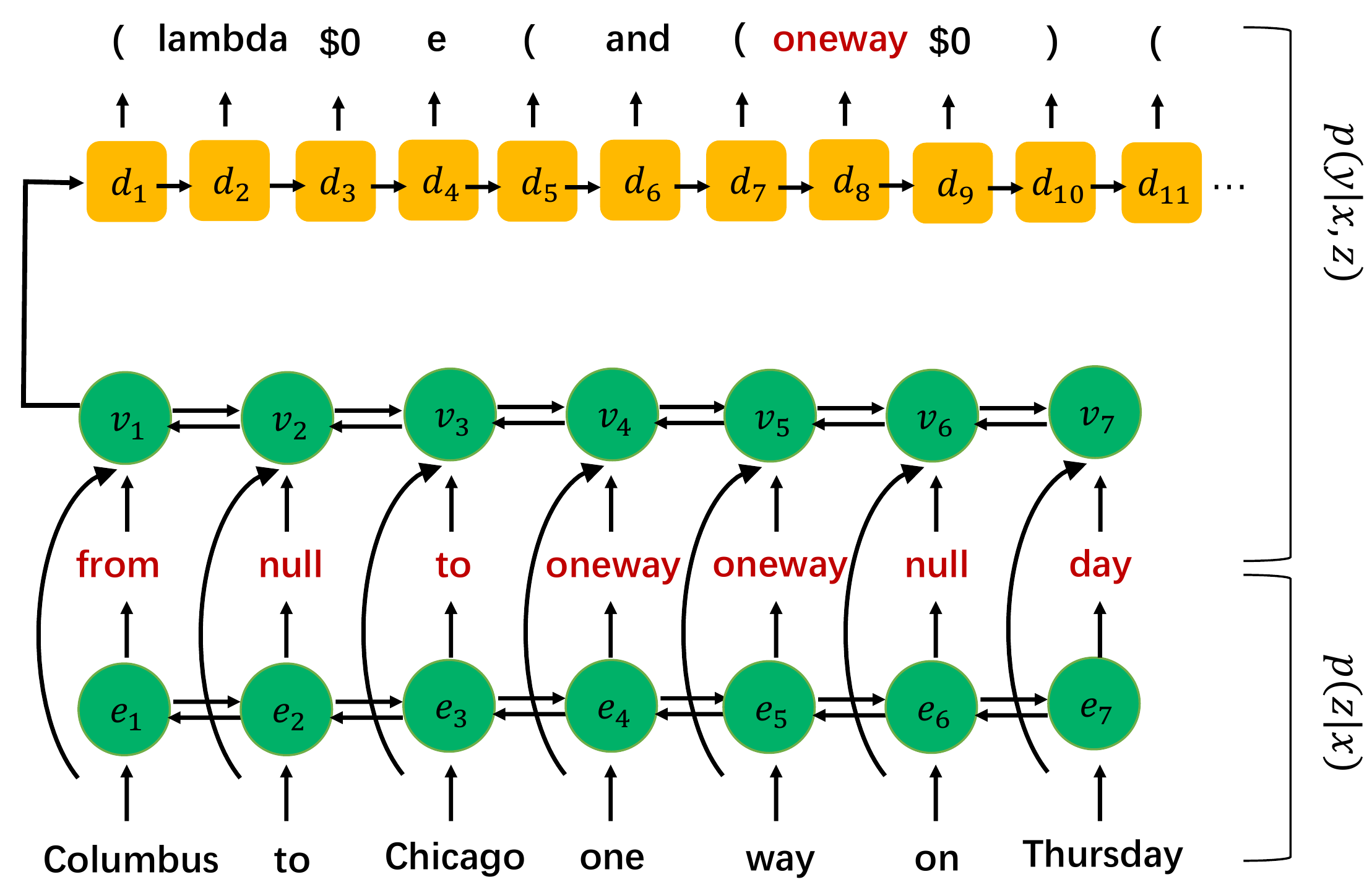}}

\caption{We first tag natural language input~$x$ with semantic symbols
  (e.g., predicates) and predict tag sequence $z$. We generate the
  final semantic representation~$y$, given $x$ and $z$ as input. }
\label{fig:method}
\vspace{-1ex}
\end{figure}

Our goal is to learn a semantic parser that takes as input a natural
language utterance \mbox{$x=x_1,x_2,...,x_n$} and predicts a meaning
representation $y=y_1,y_2,...,y_m$. We decompose the parser $p(y|x)$
into a two-stage generation process:
\begin{eqnarray}
\label{eq:two-stage}
p(y|x) = p(y|x,z)p(z|x)
\end{eqnarray}
where $z=z_1,z_2,...,z_n$ is a tag sequence for $x$. Every tag $z_t$
is a symbol in $y$ representing the meaning of $x_t$. Therefore, the
first-stage model $p(z|x)$ is essentially a tagger that tries to
predict the semantics of individual words. The second-stage model
takes word sequence~$x$ and its accompanying tag sequence~$z$ as
input, and generates the final semantic
representation~$y$. Figure~\ref{fig:method} shows the two-stage
generation process. It is important to note that tags~$z$ are
\emph{latent} and must be induced from training data, ~i.e.,~pairs of
natural language utterances and representations of their meaning.  We
discuss how the tagger is learned in
Section~\ref{sec:training}. 

\subsection{Semantic Tagging}
\label{sec:tagging}

As shown in Figure \ref{fig:method}, the tagging model $p(z|x;\theta)$
contains an encoder which transforms input sequence $x_1,x_2,...,x_n$
into a sequence of context-sensitive vector representations
$\mathbf{h}_1,\mathbf{h}_2,...,\mathbf{h}_n$.  Each word~$x_i$ is
mapped to embedding~$\mathbf{w}_i$, and the sequence of word
embeddings $\mathbf{w}_1,\mathbf{w}_2, ...,\mathbf{w}_n$ is fed to a
bi-directional recurrent neural network with long short-term memory
(LSTM) units \cite{lstm1997}. A bi-LSTM recursively computes the
hidden states at the $t$-th time step via:
\begin{eqnarray}
\overrightarrow{\mathbf{h}}_i &=& f_{\texttt{LSTM}} \quad (\overrightarrow{\mathbf{h}}_{i-1}, \mathbf{w}_i)  \\
\overleftarrow{\mathbf{h}}_i &=& f_{\texttt{LSTM}} \quad (\overleftarrow{\mathbf{h}}_{i+1}, \mathbf{w}_i) \\ 
\mathbf{h}_i &=& [\overrightarrow{\mathbf{h}}_i, \overleftarrow{\mathbf{h}}_i]
\end{eqnarray}
where $\mathbf{h}_i$ is the concatenation of vectors
$\overrightarrow{\mathbf{h}}_i$ and $\overleftarrow{\mathbf{h}}_i$, and
$f_{\texttt{LSTM}}$ refers to the LSTM function.
We feed both~$\mathbf{h}_i$ and $\mathbf{w}_i$ to the final output
layer in order to predict tags~$z$:
\begin{align}
\hspace*{-.3cm}p(z|x;\theta)& =  \prod_{i=1}^n p(z_i|x;\theta) \\
& =   \prod_{i=1}^n \operatorname{softmax}(\mathbf{W}\mathbf{h}_i + \mathbf{U}\mathbf{w}_i + \mathbf{b})
\end{align}
$\mathbf{W}$, $\mathbf{U}$, and $\mathbf{b}$ are parameters in
the output layer.

\subsection{Meaning Representation Generation}
\label{sec:mean-repr-gener}

\label{generation}

LSTM-based encoder-decoder models with an attention mechanism have
been successfully applied to a wide range of semantic parsing
benchmarks
\cite{dong-lapata-2016-language,jia-liang-2016-data,iyer-etal-2017-learning},
while Transformers have been rapidly gaining popularity for various
NLP tasks including semantic parsing
\cite{wang-etal-2020-rat,Sherborne-etal-2020-bootstrapping}.  Our approach
is model-agnostic in that it could be combined with any type of
sequence-to-sequence model; to highlight this versatility, we present
experiments with both LSTM- and Transformer-based models. We first
embed the predicted tag and word sequences, obtaining tag embeddings
$\mathbf{e}_1^g,\mathbf{e}_2^g,..,\mathbf{e}_n^g$ and word embeddings
$\mathbf{e}_1^w, \mathbf{e}_2^w, ..., \mathbf{e}_n^w$. Then, we
concatenate the two types of embeddings at each time step and feed
them to a sequence-to-sequence model:
\begin{eqnarray}
  \mathbf{u}_t & = &  [\mathbf{e}_t^g, \mathbf{e}_t^w] \\
  y & = & f_{\texttt{seq2seq}}(\mathbf{u})
\end{eqnarray}
where $[\cdot,\cdot]$ denotes vector concatenation and
$f_{\texttt{seq2seq}}$ denotes a sequence-to-sequence model variant
(LSTM- or Transformer-based in our case) that takes a sequence of
vector representations as input and ultimately generates a logical
form.  Tag embeddings are shared with the embeddings used in the
decoder. Therefore, the only adaptation we make to the baseline model
is replace the original word embeddings with tag-augmented input.

\begin{table}[t]
\begin{small}
\begin{tabular}{@{}l@{~~}p{7.2cm}@{}}
\hline 
 \multicolumn{2}{c}{$\lambda$-calculus} \\ \hline
 $x:$ & {Columbus to Chicago one way on Thursday} \\
 $z:$ & {Columbus}/\texttt{\textbf{from}} {to}/\texttt{\textbf{null}} {Chicago}/{\bf to}  {one}/\texttt{\textbf{oneway}}  {way}/\texttt{\textbf{oneway}}  {on}/{\bf null} {Thursday}/\texttt{\textbf{day}} \\
 $s:$& \texttt{\textbf{oneway}, \textbf{from}, \textbf{to}, \textbf{day}}\\
 $y:$& \cmtt{( lambda \$0 e ( and ( \texttt{\textbf{oneway}} \$0 ) ( \texttt{\textbf{from}} \$0 columbus:ci ) ( \texttt{\textbf{to}} \$0 chicago:ci ) ( \texttt{\textbf{day}} \$0 thursday:da ) ) )} \\ 
  \hline 
\multicolumn{2}{c}{} \\\hline
 \multicolumn{2}{c}{SQL} \\ \hline
 $x:$& {What is the area of Washington} \\
 $z:$ &   {What}/\texttt{\textbf{null}} {is}/\texttt{\textbf{null}}  {the}/\texttt{\textbf{null}} {area}/\texttt{\textbf{area}} {of}/\texttt{\textbf{null}} {Washington}/\texttt{\textbf{state\_name}} \\
 $s:$& \texttt{\textbf{area, state\_name}}\\
 $y:$ & \cmtt{select \texttt{\textbf{area}} from state where \texttt{\textbf{state\_name}} = "washington"}  \\ 
  \hline

\end{tabular}

\end{small}
\caption{
  Utterances $x$, their
  meaning representations $y$, symbol sets~$s$, and predicted
  word/\texttt{\textbf{tag}} sequences $z$.} \label{table:example} 

\vspace{-0.1in}
\end{table}

\section{Model Training}
\label{sec:training}

Our proposed approach combines a semantic tagger with a
sequence-to-sequence model. The tagger learning problem is challenging
since $z$~is unobserved. In this section, we explain how the tagger
and the overall model are trained.

\subsection{Tagger Learning}
We learn a tagger~$p(z|x;\theta)$ from training data consisting of
pairs of natural language utterances $ x = x_1,x_2,...,x_n$ and symbol
sets $s = \{s_1,s_2,...,s_l\}$ (with~$s_j \in y$).  The symbol set
contains atomic semantic units such as \mbox{$\lambda$-calculus}
predicates and SQL column names.  Table~\ref{table:example} presents
examples of symbol sets for these two formalisms. As can be seen,
symbols have close ties to utterances, there is often a correspondence
between them and individual words or phrases. It is therefore natural
to predict this (basic) part of a meaning representation via a
tagger. To bridge the gap between the tag sequence we intend to
predict and the symbol set we have as supervision, we introduce latent
variable $a = a_1, a_2, ..., a_n$ where $a_j$ denotes the index of a
word aligned to $s_j$. We add $(n-l)$ $\mathsf{null}$ symbols to
target set $s = \{s_1,s_2,...,s_l,s_{l+1},...,s_n\}$ because~$n$ is
typically larger than~$l$, and we allow the tagger to output
$\mathsf{null}$ for some words.

\paragraph{Entity Linking} For some symbols, it is rather
straightforward to determine the corresponding alignments based on the
results of entity linking, a critical subtask in semantic parsing
which is generally treated as a preprocessing step
\cite{dong-lapata-2016-language,jia-liang-2016-data}. We thus define
the following two rules to automatically align symbols to words in an
utterance based on entity linking: (1)~for \mbox{$\lambda$-calculus}
expressions, if a predicate takes only one entity as an argument
(e.g., \cmtt{day \$0 thursday:da}) and this entity can be linked to a
word or phrase in the utterance, we assume there is an alignment
between them (e.g.,~\cmtt{day} aligns to \textsl{Thursday}); (2)~for
SQL expressions, if the entity in a filter clause
(e.g.,~\cmtt{state\_name = "washington"}) can be linked to an
expression in the utterance, again we align the column
(e.g.,~\cmtt{state\_name}) to the linguistic expression
(e.g.,~\textsl{Washington}). Both rules capture the intuition that
some semantic symbols are implied by corresponding entities without
being explicitly verbalized. As shown in
\mbox{Table~\ref{table:example}}, there is no linguistic expression in
the utterance ``\emph{What is the area of Washington}'' which
corresponds to the logical expression of \cmtt{state\_name}, instead
\cmtt{state\_name} is implied by the entity \cmtt{washington}.

\paragraph{Expectation-Maximization} Besides entity linking, there
remain symbols without alignments, such as unary predicates (e.g.,
\cmtt{oneway \$0}). For these, we use an EM-style algorithm which
iteratively infers latent alignments~$a$ and uses them to update the
tagger. A hard-EM algorithm that predicts the most probable~$a$ seems
reasonable as in most cases there is a single correct
alignment. However, we find that hard-EM renders training unstable and
prone to overfitting to incorrect alignments. We instead warm up the
training with a soft-EM algorithm first and switch to hard-EM later
on.
Without loss of generality, we describe the algorithm for all symbols
including those that could be aligned via (entity linking)
rules. Specifically, we model the generation of~$s$ as follows:
\setlength{\abovedisplayskip}{10pt}
\setlength{\belowdisplayskip}{10pt}
\setlength{\abovedisplayshortskip}{0pt}
\setlength{\belowdisplayshortskip}{0pt}
 \begin{eqnarray}
p(s|x;\theta)&\hspace*{-.2cm}=\hspace*{-.2cm}& \sum_{a} p(a|x)p(s|x,a;\theta) \nonumber \\
     &\hspace*{-.2cm}=\hspace*{-.2cm}& \sum_{a} p(a|x) \prod_{j=1}^n p(z_{a_j}=s_j|x;\theta)~ \quad 
\end{eqnarray}
where $p(a|x)$ is a uniform prior over $a$ and
\mbox{$p(z_{a_j}=s_j|x;\theta)$} is the tagger model above.
We could constrain the alignment from words to symbols to be injective
(as this would more faithfully capture the complex dependencies
between them). Unfortunately, this renders posterior inference on~$a$
intractable.  Instead, we model the alignment of each symbol
independently as:
\begin{eqnarray}
\hspace*{-.3cm}p(s|x;\theta)&\hspace*{-.3cm}=\hspace*{-.3cm}& \sum_{a} \prod_{j=1}^n p(a_j|x) \prod_{j=1}^n p(z_{a_j}=s_j|x;\theta) \nonumber \\
 &\hspace*{-.3cm}=\hspace*{-.3cm}& \prod_{j=1}^n \sum_{a_j} p(a_j|x) p(z_{a_j}=s_j|x;\theta)~ 
\end{eqnarray}
Under this assumption, we are able to exactly compute the posterior probability of each alignment~$a_j$:
\begin{eqnarray}
\pi_{ij}(\theta) & \hspace*{-.2cm}=\hspace*{-.2cm} & p(a_j=i|x, s; \theta) \nonumber \\
& \hspace*{-.2cm}=\hspace*{-.2cm} & \frac{p(a_j=i|x) p(z_i=s_j|x;\theta)}{\sum_{\tilde{i}=1}^n p(a_j=\tilde{i}|x) p(z_{\tilde{i}}=s_j|x;\theta)} \quad \quad
\end{eqnarray}
Note that we manually set the value of $\pi_{ij}(\theta)$ if~$a_j$
can be induced in advance via entity linking. At the $t$-th iteration,
we first use the present tagger~$p(z|x;\theta^t)$ to compute
$\pi_{ij}(\theta^t)$, the likelihood of aligning~symbol $s_j$ to
word~$x_i$. For soft-EM, these assignments are then directly used to
train the tagger with the following objective:
\begin{eqnarray}
 \mathcal{J}_t(\theta) & \hspace*{-.2cm}=\hspace*{-.2cm} & \sum_{i=1}^n \sum_{j=1}^n \pi_{ij}(\theta^t) \log p(z_i=s_j|x; \theta) ~~ \quad \\
 \theta^{t+1} & \hspace*{-.2cm}=\hspace*{-.2cm} & \argmax_\theta~ \mathcal{J}_t(\theta)
\end{eqnarray}
\indent For hard-EM, one could exploit $\pi_{ij}(\theta^t)$ to induce the most
probable alignment for each symbol. However, there are cases where a
symbol is aligned to multiple words, e.g., when the same word occurs
multiple times in an utterance or when a symbol is aligned to a
phrase.  To deal with such cases, we induce a hard-version of the
posterior probability $\tilde{\pi}_{ij}(\theta^t)$ in the following
way:
\begin{eqnarray}
\tilde{\pi}_{ij}(\theta^t)  &=& 
\begin{cases}
    1  &  \text{if } \pi_{ij}(\theta^t)~ >~ \beta \\
    0  &  \text{otherwise}
 \end{cases} \quad \\
 && \quad \quad \quad \quad (1 \leq j \leq l) \nonumber \\
 \tilde{\pi}_{ij}(\theta^t) &=& \frac{1 - \sum_{k=1}^l \tilde{\pi}_{ik}}{n-l} \label{null_pos}\\ 
 && \quad \quad \quad (l+1 \leq j \leq n) \nonumber
\end{eqnarray}
where $\beta$ is a threshold used to discretize the soft alignment
distributions. Reshaping the posteriors in this manner allows a symbol
to be aligned to multiple words while removing noisy incorrect
alignments. Equation \eqref{null_pos} ensures that the sum of
posteriors corresponding to a word is one, in the hope of encouraging
the predicted tag sequence distribution to be as close to a normal tag
sequence distribution as possible. We replace $\pi_{ij}(\theta^t)$ in
$\mathcal{J}(\theta|\theta^t)$ with $\tilde{\pi}_{ij}(\theta^t)$ as
the training objective to perform hard-EM updates:
\begin{eqnarray}
 \tilde{\mathcal{J}}_t(\theta) & \hspace*{-.2cm}=\hspace*{-.2cm} & \sum_{i=1}^n \sum_{j=1}^n \tilde{\pi}_{ij}(\theta^t) \log p(z_i=s_j|x; \theta) ~~ \quad \\
 \theta^{t+1} & \hspace*{-.2cm}=\hspace*{-.2cm} & \argmax_\theta~  \tilde{\mathcal{J}}_t(\theta)
\end{eqnarray}
Our training procedure is shown in
Algorithm~\ref{algorithm:tagger}. Note that in each EM iteration, we
use objective $ \mathcal{J}_t(\theta)$ or
$\tilde{\mathcal{J}}_t(\theta)$ to compute the gradient and update
parameters once rather than maximizing the objective function.

\begin{algorithm}[t]
\SetAlgoLined
\KwIn{Dataset $\mathcal{D}$ where each example is a question~$x$
  paired with symbol set~$s$. Number of soft-EM updates
  $T_s$. Number of overall updates $T$.}  \KwOut{Tagger model
  parameters~$\theta^{T+1}$} \ Initialize tagger parameters $\theta^1$
randomly\; \For{$t=1,...,T$}{
  sample an example ($x$, $s$) \\
  \eIf{$t < T_s$}{ \tcc{do soft-EM update}
    Compute $\pi_{ij}(\theta^t)$  \\
    $\theta^{t+1} \leftarrow \operatorname{Optimizer}(\theta^t, \nabla_{\theta_t}\mathcal{J}_t(\theta_t)) $\\
   }{
   \tcc{do hard-EM update}
   Compute $\tilde{\pi}_{ij}(\theta^t)$ \\
    $\theta^{t+1} \leftarrow \operatorname{Optimizer}(\theta^t,\nabla_{\theta_t}\tilde{\mathcal{J}}_t(\theta_t)) $\\
  }
 }
 \KwRet{$\theta^{T+1}$}
 \caption{Training the tagger}
 \label{algorithm:tagger}

\end{algorithm}

\subsection{Parser Learning}
Learning a semantic parser in our setting is straightforward. After
training the tagger, we run it over the examples in the training data
and obtain tag sequence~$\hat{z}$ for each pair of utterance $x$ and
meaning representation~$y$.
\begin{eqnarray}
\hat{z} = \argmax_z p(z|x;\theta)
\end{eqnarray}
Then, we maximize the likelihood of generating $y$ given $x$ and $\hat{z}$:
\begin{eqnarray}
\hat{\theta} = \argmax_\theta \log p(y|x,\hat{z};\theta)
\end{eqnarray}

 

\section{Experimental Setup}

\paragraph{Datasets} Our experiments evaluate the proposed framework
on compositional generalization.  We present results on query-based
splits for three widely used semantic parsing benchmarks, namely
\textsc{Atis} \cite{atis-Dahl:1994:ESA:1075812.1075823},
\textsc{GeoQuery} \cite{Zelle1996}, and \textsc{WikiSQL}
\cite{zhongSeq2SQL2017}. For \textsc{GeoQuery} (880 language queries
to a database of U.S. geography) and \textsc{Atis} (5,410 queries to a
flight booking system) meaning representations are in
\mbox{$\lambda$-calculus} and SQL. We adopt the split released by
\citet{finegan-dollak-etal-2018-improving} for SQL. We create query-based splits for
\mbox{$\lambda$-calculus}, as we use the preprocessed versions provided in
\citet{dong-lapata-2018-coarse}, where natural language expressions
are lowercased and stemmed with NLTK \cite{bird2009natural}, and
entity mentions are replaced by numbered markers.

\textsc{WikiSQL} is a large-scale semantic parsing dataset released
more recently \cite{zhongSeq2SQL2017}. It is used as a testbed for
generating an SQL query given a natural language question and table
schema (i.e.,~table column names) without using the content values of
tables. Since SQL queries in most examples are simple and only contain
one filtering condition, we use a subset (16,835 training examples,
2,602~validation examples, and~4,915 test examples) containing queries
with more than one filtering condition. These examples are more
compositional and better suited to evaluating compositional
generalization.

\paragraph{Comparison Models}
On \textsc{Atis} and \textsc{GeoQuery} we trained two baseline
sequence-to-sequence models which we implemented using LSTMs and
Transformers as the base units (see Section~\ref{generation}).  To
examine whether our results carry over to pretrained contextual
representations, we report experiments with an LSTM model enhanced
with RoBERTa \cite{liu2019roberta}.  We also compare against two
related approaches.  The first is \textsc{GECA}
\cite{andreas-2020-good}, a recently proposed data augmentation method
aimed at providing a compositional inductive bias into
sequence-to-sequence models.  The second is Attention Supervision
introduced in \citet{Oren:ea:2020}.  They encourage generalization by
supervising the decoder attention with pre-computed token
alignments. We use the alignments induced by our tagger instead of an
off-the-shelf word aligner adopted in their paper.


For \textsc{WikiSQL}, our baseline model follows the
\textsc{coarse2fine} approach put forward in
\citet{dong-lapata-2018-coarse} which is well suited to the formulaic
nature of the queries, takes the table schema into account, and
performs on par with some more sophisticated models
\cite{mccann2018natural, yu-etal-2018-typesql}.  They predict
\cmtt{select} and \cmtt{where} SQL clauses separately (all queries in
\textsc{WikiSQL} follow the same format, i.e.,~"\cmtt{SELECT agg\_op
  agg\_col where (cond\_col cond\_op cond AND)}...", which is a small
subset of the SQL syntax).  The \cmtt{select} clause is predicted via
two independent classifiers, while the \cmtt{where} clause is
generated via a sequence model with a sketch as an intermediate
outcome. Their encoder augments question representations with table
information by computing attention over table column vectors and
deriving a context vector to summarize the relevant columns for each
word.


Our tagger uses \textsc{coarse2fine}'s table-aware encoder to predict
tags. Our parser diverges slightly from their model: while for each
word the context vector is originally computed by the attention
mechanism, we replace it with the column vector specified by the
corresponding tag.

\paragraph{Configuration}
We implemented the base semantic parsers (\textsc{LSTM} and
\textsc{Transformer}) with fairseq \cite{ott2019fairseq}. As far as
\textsc{GECA} is concerned, we have a different setting from
\citet{andreas-2020-good}: we use the preprocessed versions provided
by \citet{dong-lapata-2018-coarse} for \textsc{Atis} and
\textsc{GeoQuery}, while they report experiments on \textsc{GeoQuery}
only, with different preprocessing. We used their open-sourced code to
generate synthetic data for our setting in order to make experiments
comparable.  For \textsc{coarse2fine} \cite{dong-lapata-2018-coarse},
we used the code released by the authors.

Hyperparameters for the semantic taggers were validated on the
development split of \textsc{Atis} and were directly copied for
\textsc{GeoQuery} because of its small size. Dimensions of hidden
vectors and word embeddings were selected from \{150, 200, 250,
300\}. The number of layers was selected from \mbox{\{1, 2\}}. Batch
size was set to 20 and the overall update step was set to 20,000. The
number of steps for soft-EM updates was selected from \{5,000, 7,000,
10,000, 13,000\}. The threshold $\beta$ used in hard-EM was selected
from \{0.20, 0.23, 0.26, 0.29, 0.32, 0.35\}. We used the Adam
optimizer \cite{KingmaB14} to train the models and the learning rate
was selected from \{0.0001, 0.0003, 0.001\}. Our semantic parsers used
the same hyperparameters as the base models except for some necessary
changes to incorporate tag inputs. For models using RoBERTa, we first
freeze RoBERTa and train the model for some steps, and then resume
fine-tuning.

\paragraph{Evaluation}
We use exact-match accuracy as our evaluation metric, namely the
percentage of examples that are correctly parsed to their gold
standard meaning representations. For \textsc{WikiSQL}, we also
execute generated SQL queries on their corresponding tables, and
report execution accuracy which is defined as the proportion of
correct answers.

\begin{table}[t]
\centering
\resizebox{\linewidth}{!}{
\begin{tabular}{@{}l|c@{~~}c|c@{~~}c@{}}
  \thickhline
 \multirow{2}{*}{{\bf Method}}  & \multicolumn{2}{c|}{{\bf $\lambda$-calculus}} & \multicolumn{2}{c}{{\bf SQL}} \\

&  \textsc{Geo}& \textsc{Atis} &  \textsc{Geo} & \textsc{Atis} \\

  \thickhline
  \textsc{GECA} & 48.1 & 51.6 & 52.1 & 24.0   \\\hline\hline
  \textsc{Transformer} & 39.8 & 51.2  & 53.9 & 23.0 \\
  \textsc{Transformer + AS} & 43.4 & 53.3  & 58.6 & 22.0 \\ 
  \textsc{Transformer} + ST & 44.0 & 53.0 & 61.9 & 28.6  \\\hline\hline
  \textsc{LSTM} & 49.8 & 56.2 & 48.5 & 28.0  \\
     \textsc{LSTM + AS } & 53.6 & 59.7 & 46.9 & 28.7  \\
  \textsc{LSTM} + ST  & 52.1 & 62.1 & 63.6 & {29.1}   \\ \hline\hline
  
  \textsc{RoBERTa} & 54.4 & 57.5 & 58.8 & 28.6  \\
     \textsc{RoBERTa + AS } & 56.3 & 59.9 & 59.3 & 28.4  \\
  \textsc{RoBERTa} + ST  & {57.5} & {63.7} & {69.6} & 27.7   \\

   \thickhline
\end{tabular}
}
\caption{Exact-match accuracy on \textsc{GeoQuery} and
  \textsc{Atis}; results averaged over 5 random seeds; ST stands
  for semantic tagging; AS is attention supervision.}  

\label{table:results}
\end{table}

\section{Results}
\label{sec:results-analysis}

\begin{table}[t]
\centering
\resizebox{\linewidth}{!}{
\begin{tabular}{l c c c}
	\thickhline 
\textbf{Method}                   & \textbf{Acc} & \textbf{Exe} &
\texttt{\textbf{where}} \\
	 \thickhline
	\textsc{coarse2fine}                & 58.0       & 68.2   & 71.3                           \\
	\textsc{coarse2fine} + AS        & 58.8             & 69.2   & 72.8                     \\
	\textsc{coarse2fine} + ST        & 60.6             & 71.3   & 75.0                     \\
	 \thickhline
\end{tabular}
}
\vspace{-0.5ex}

\caption{Evaluation results on a 
  \textsc{WikiSQL} subset. \textbf{Acc}:  exact-match
  accuracy; \textbf{Exe}:  execution accuracy;
  \texttt{\textbf{where}}: accuracy of predicting \cmtt{where}
  clauses.} 
\label{tbl:results:wikisql}
\vspace{-0.5ex}

\end{table}

\begin{table*}[t]
\centering
\begin{tabular}{lr@{}r@{}rr@{}r@{}rr@{}r@{}rr@{}r@{}r}
  \thickhline
 \multirow{2}{*}{{\bf Method}}  & \multicolumn{6}{c}{{\bf $\lambda$-calculus}} & \multicolumn{6}{c}{{\bf SQL}} \\

&  \multicolumn{3}{c}{\textsc{Geo}}& \multicolumn{3}{c}{\textsc{Atis}} &  \multicolumn{3}{c}{\textsc{Geo}} & \multicolumn{3}{c}{\textsc{Atis}} \\

  \thickhline
 
  \textsc{LSTM} & 16.1~/& ~10.9~/& ~23.2 & 13.7~/& ~9.8~/& ~20.1 & 14.1~/
  &~5.6~/ & ~31.8 & 21.3~/ & ~26.5~/ & ~23.7  \\
  \textsc{LSTM} + ST & 16.5~/ & ~9.7~/ & ~21.7 & 13.0~/ & ~8.9~/ &
  ~15.9 & 19.0~/ & ~6.7~/ & ~10.5 & 22.1~/ & ~24.9~/ & ~23.9 \\

  \textsc{RoBERTa} + ST  & 13.0~/ & ~9.6~/ & ~19.7 & 12.9~/ & ~6.9~/ &
  ~16.4 & 12.7~/ & ~5.7~/ & ~11.8 & 22.6~/ & ~16.6~/ & ~32.8 \\

   \thickhline
\end{tabular}
\caption{Breakdown of different types of error.  In each cell, left  shows the
  proportion of predicting correct semantic symbols but incorrect
  queries;  middle is the proportion of predicting a subset of
  correct symbols (i.e.,~missing some semantic symbols);
  right is the proportion of predicting  symbols which
  do not exist in gold queries. }  
\vspace{-0.1in}
\label{table:analysis}
\end{table*}


\paragraph{Does Tagging Help  Parsing?}
Table~\ref{table:results} summarizes our results on \textsc{Atis} and
\textsc{GeoQuery}. On both datasets, we observe that the proposed
tagger ($+$ ST) boosts the performance of the base model
(\textsc{Transformer}, \textsc{LSTM}) for both
\mbox{$\lambda$-calculus} and SQL. The \textsc{LSTM} is generally
superior to \textsc{Transformer} except on SQL
\textsc{GeoQuery}. Enhancing the LSTM with pretrained contextual
representations (see the last block in the table) generally increases
accuracy, yet our semantic tagger brings improvements on top of
\textsc{RoBERTa} (with the exception of SQL \textsc{Atis}). This
points to the generality of our approach which benefits neural parsers
with different architectures trained on distinct semantic
representations. Gains are particularly significant on \textsc{Atis}
with \mbox{$\lambda$-calculus} (we observe an absolute improvement of
6.2~points over \textsc{RoBERTa}) and \textsc{GeoQuery} with SQL (with
10.8~points absolute improvement over \textsc{RoBERTa}).

In some settings, attention supervision (+AS) also achieves
improvements over baseline sequence models, but these are inconsistent
and sometimes it even slightly hurts performance. We find that
attention supervision is sensitive to the weight hyperparameter that
controls the strength of attention loss and requires careful tuning to
achieve good performance. We conjecture that the soft attention
mechanism (even with proper supervision signals) is still sensitive to
irrelevant context changes and prone to errors in cases requiring
compositional generalization. The \textsc{LSTM}+ST model achieves
better accuracy than \textsc{GECA} which adopts a data augmentation
strategy to train a LSTM-based sequence-to-sequence model for
compositional generalization. We incorporate a similar inductive bias
into the parser, but in an orthogonal way.

Results on \textsc{WikiSQL} are shown in
Table~\ref{tbl:results:wikisql}. Semantic tagging boosts
\textsc{coarse2fine} in terms of exact match and execution
accuracy. In particular, it improves the prediction of \cmtt{where}
clauses, by a 4.3\% absolute margin.  We would not expect semantic
tagging to benefit any other parts of the generation of the SQL query,
since only \cmtt{where} clauses are decoded sequentially in the
\textsc{coarse2fine} model. Gains in the generation of \cmtt{where}
clauses translate to improvements in overall accuracy. Attention
supervision (+AS) also improves generalization but falls behind our
semantic tagger.

\paragraph{Do Meaning Representations Matter?}
Improvements of our semantic tagger on \textsc{Atis} with SQL and
\textsc{GeoQuery} with \mbox{$\lambda$-calculus} are less dramatic
compared to \textsc{Atis} with \mbox{$\lambda$-calculus} and
\textsc{GeoQuery} with SQL.  Upon closer inspection, we find that
\textsc{Atis} SQL queries typically include many bridging columns that
are used to join two tables. This arises from the complex database
structure in \textsc{Atis}: there are 32 tables in total and each
query involves~6.4 tables on average. These bridging columns are
SQL-specific and generally do not align with any linguistic
expressions, so we cannot improve their prediction via semantic
tagging. A prerequisite for semantic tagging is that there exist
alignments between language expressions and atomic semantic
symbols. We could restrict the semantic tagger to only predicting
symbols which align to linguistic expressions and leave the generation
of other symbols to the second stage. However, how to automatically
select appropriate symbols as semantic tags is an avenue for future
work.

On \textsc{GeoQuery} with \mbox{$\lambda$-calculus}, the semantic
tagger performs extremely well, achieving 86.2\% accuracy in
predicting semantic symbols, but the final accuracy in predicting
queries is only 52.1\% (\textsc{LSTM}+ST). Although semantic tagging
can help generalize to utterances where seen syntactic structure and
concept words are combined in an unseen way (e.g., \textsl{Monkeys
  like bananas} generalizes to \textsl{Cats like fish}), it fails to
generalize to utterances with unseen syntactic structure (e.g.,
\textsl{Monkeys like bananas} generalizes to \textsl{Cats like fish
  that like water}). Handling utterances with unseen composition of
seen syntactic components is yet another generalization challenge for
modern semantic parsers.

\paragraph{Where do Gains Come from?}
Our approach transfers much of the prediction of semantic symbols from
the sequence-to-sequence model to the tagger; it does this by
replacing the attention mechanism, which learns to attend to
\emph{specific} parts of an utterance, with per-word tagging which
considers \emph{all} parts of an utterance. We hypothesize that this
architecture can better exploit source information to predict
individual semantic symbols. To test this hypothesis, we analyzed
errors in the predictions of the \textsc{LSTM} model with and without
the proposed semantic tagger, and classified them into three
types. The first type predicts incorrect queries but correct semantic
symbols. The second type predicts only a subset of correct semantic
symbols, thus omitting some semantic symbols. The third type predicts
wrong semantic symbols that do not exist in gold queries. As shown in
Table~\ref{table:analysis}, semantic tagging mainly reduces the errors
of predicting wrong semantic symbols, while in some cases it can lead
to a modest increase in the first type of errors. Overall, semantic
tagging improves the prediction of individual semantic symbols even
though this improvement does not always translate into more accurate
queries.

\section{Conclusions} 
We presented a two-stage decoding framework, aiming to improve
compositional generalization in neural semantic parsing. Central to
our approach is a semantic tagger which labels the input with semantic
symbols representing the meaning of individual words. A neural
sequence-to-sequence parsing model consider the input utterance
and the predicted tag sequence to generate the final meaning
representation. Our framework can be combined with different neural
models and semantic formalisms and demonstrates superior performance
to related compositional generalization approaches
\cite{andreas-2020-good,Oren:ea:2020}.  In the future, we would like
to extend our approach to learning syntactic generalizations.

\paragraph{Acknowledgments} We thank the anonymous reviewers for their
feedback. We gratefully acknowledge the support of the European
Research Council (award number 681760).

\bibliography{custom}
\bibliographystyle{acl_natbib}

\end{document}